\pdfoutput=1

\documentclass[11pt]{article}

\usepackage[]{EMNLP2023}

\usepackage{times}
\usepackage{latexsym}

\usepackage[T1]{fontenc}

\usepackage[utf8]{inputenc}

\usepackage{microtype}

\usepackage{inconsolata}

\usepackage{microtype}
\usepackage{graphicx}
\usepackage{subfigure}
\usepackage{amsmath}
\usepackage{amsfonts}
\usepackage{multirow}
\usepackage{color}
\usepackage{amssymb}

\usepackage{booktabs}
\usepackage{arydshln}

\title{Self-Adaptive Reconstruction with Contrastive Learning for Unsupervised Sentence Embeddings}

\author{Junlong Liu, Xichen Shang, Huawen Feng, Junhao Zheng, Qianli Ma\footnotemark[1] \\
  School of Computer Science and Engineering, \\ 
  South China University of Technology, Guangzhou, China \\
  \texttt{junlongliucs@foxmail.com} \\ \texttt{qianlima@scut.edu.cn\footnotemark[1]}}

\begin{document}
\maketitle

\renewcommand{\thefootnote}{\fnsymbol{footnote}}
\footnotetext[1]{Corresponding author}

\renewcommand{\thefootnote}{\arabic{footnote}}

\begin{abstract}

Unsupervised sentence embeddings task aims to convert sentences to semantic vector representations. Most previous works directly use
the sentence representations derived from pretrained language models. However, due to the token bias in pretrained language models, the models can not capture the fine-grained semantics in sentences, which leads to poor predictions. To address this issue, we propose a novel \textbf{S}elf-\textbf{A}daptive \textbf{R}econstruction \textbf{C}ontrastive \textbf{S}entence \textbf{E}mbeddings (\textbf{SARCSE}) framework, which reconstructs all tokens in sentences with an AutoEncoder to help the model to preserve more fine-grained semantics during tokens aggregating. In addition, we proposed a self-adaptive reconstruction loss to alleviate the token bias towards frequency. Experimental results show that SARCSE gains significant improvements compared with the strong baseline SimCSE on the 7 STS tasks.

\end{abstract}

\section{Introduction}

The goal of unsupervised sentence embeddings is to learn semantic sentence representations. It could be widely used in downstream tasks. The sentence embeddings are generally directly derived from pretrained language models (PLMs) like BERT \citep{devlin-etal-2019-bert} and RoBERTa \citep{liu2019roberta} in previous works. To alleviate the problem of anisotropy \citep{li-etal-2020-sentence} in PLMs, recently, \citet{gao-etal-2021-simcse} proposes SimCSE based on contrastive learning, which obtained positive pairs from multiple dropouts \citep{JMLR:v15:srivastava14a} and negative pairs from the sentences in the same mini-batches. Finally, it uses the representations of token $[CLS]$ as the sentence representations, then pulls the positive pairs closer and pushes the negative pairs further away using the InfoNCE loss.

\begin{figure}[t!]
    \centering
    \includegraphics[width=\linewidth]{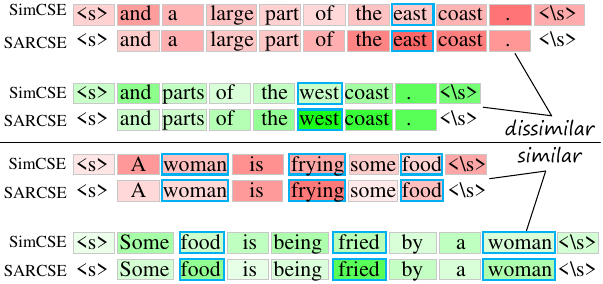}
    \caption{
    Two examples extracted from the corpus. The tokens with blue borders are the keywords in the two sentences. The deeper color of tokens means the greater importance in sentence embeddings for the two models. The importance of tokens in SimCSE is obtained from the self-attention aggregation weights of <s>. And the importance of tokens in SARCSE is obtained from the reconstruction loss of AutoEncoder. The deeper color means lower reconstruction loss. The tokens <s> and <$\backslash$s> have no loss because we do not reconstruct them. SARCSE pays more attention to fine-grained differences, but SimCSE does not, causing SimCSE to make a wrong similarity prediction between two sentences.
    }
    \label{fig:example}
\end{figure}

However, two problems degrade the performance of models which based on SimCSE. One is that the token $[CLS]$ ignores some fine-grained semantics during tokens aggregation. This results in the misjudgment of sentence similarity, especially when there is only a fine-grained difference between two sentences. The other problem is that the tokens with different frequencies non-uniformly distribute in representation space in PLMs, termed as token bias towards frequency \citep{jiang2022promptbert}, which degrades the performance of SimCSE. For instance, we show a dissimilar and a similar example in Figure~\ref{fig:example}. SARCSE makes better predictions than SimCSE in both cases. In the dissimilar sentence pair, the most significant difference between these two sentences are the words "east" and "west". And it leads to absolute opposite semantics in two sentences. However, SimCSE pays more attention to the high-frequency tokens (e.g., "and", ".", "<s>") rather than the determinative semantic keywords "east" and "west" by observing the self-attention aggregation weights of token <s>. Moreover, in the similar sentence pair, SARCSE still pays more attention to key semantic tokens (e.g., "woman", "frying" and "food"), but SimCSE pays more attention to some inessential tokens (e.g., "is", "A"/"a" and "Some"). This difference increases incorrect predictions in SimCSE.

Given the above-mentioned situation, we propose a novel \textbf{S}elf-\textbf{A}daptive \textbf{R}econstruction \textbf{C}ontrastive \textbf{S}entence \textbf{E}mbeddings (\textbf{SARCSE}) framework, which can identify the subtle differences between two sentences and mitigate token bias. Specifically, we use an AutoEncoder after the PLMs to reconstruct all tokens in sentences to force the model to preserve the fine-grained semantics as much as possible. Inspired by \citet{jiang2022promptbert} and \citet{10.1145/3477495.3531823}, to reduce the impact of token bias, we propose a self-adaptive reconstruction loss based on token frequency. It is worth noting that SARCSE is an upgrade to the sentence encoder, which is plug-and-play for other strong baselines based on data augmentation. Experimental results on 7 STS tasks demonstrate the effectiveness of SARCSE compared with SimCSE.

\section{Method}

\begin{figure}[t!]
    \centering
    \includegraphics[width=\linewidth]{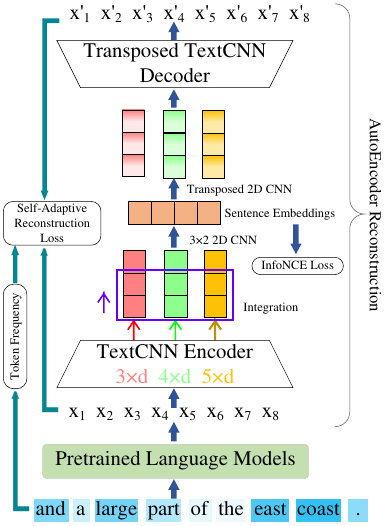}
    \caption{
    The overall architecture of SARCSE. The deeper blue on tokens means the lower token frequency, which shows most high-frequency tokens have no determinative semantics. First, the tokens of sentences are input into the pretrained language models to get token representations. Then, the multi-scale representations are obtained using TextCNN Encoder with different convolution kernels. Integrating them with a CNN will get the sentence embeddings. Finally, we use the transposed CNN and TextCNN to reconstruct the sequence of tokens.
    }
    \label{fig:method}
\end{figure}

In this section, we mainly describe SARCSE, which reconstructs tokens with the self-adaptive reconstruction loss. The structure of SARCSE is shown in Figure~\ref{fig:method}.

\subsection{Token Representations}

Given a sentence $S=\{w_1, w_2, \ldots, w_N\}$ consisting of $N$ tokens, we feed $S$ into Roberta \citep{liu2019roberta}. 
Specifically, we use the token representations except for the <s> and <$\backslash$s> in each sentence. Hence, the sentence with $N$ tokens can be represented as:
\begin{equation}
    X = \left\{ x_1,  x_2, \ldots, x_N\right\}
\end{equation}
where $x_i \in \mathbb{R}^{d}$ and $d$ is the hidden size of RoBERTa.

Following \citet{gao-etal-2021-simcse}, we input the sentences twice to get the positive samples in contrastive learning by a random mask for dropout:
\begin{equation}
    X^+ = \left\{ x_1^+,  x_2^+, \ldots, x_N^+\right\}
\end{equation}

\subsection{Reconstruction with AutoEncoder}
In order to preserve more fine-grained semantics in sentences, we use an AutoEncoder to reconstruct the input sentences.

Firstly, we input the token representation $X$ into the encoder based on TextCNN \citep{kim-2014-convolutional}. Specifically, we encode the tokens using convolution kernels of different sizes. And each kind of convolution kernel uses the same input $X$:
\begin{equation}
    H_{ks} = TextCNN_{ks}(X), ks=3,4,5
\end{equation}
where the token window (or kernel size) of each TextCNN is $(ks \times d)$, $H_{ks} \in \mathbb{R}^{co_t}$ and $co_t$ is the number of output channels of TextCNN.

Given the multi-scale context representations, to jointly consider them, we use a Convolutional Neural Network (CNN) to integrate the representations of different scales:
\begin{equation}
    Z = CNN([H_3;H_4;H_5])
\end{equation}
where the kernel size of CNN is $(3 \times 2)$, $Z \in \mathbb{R}^{co_c*(co_t - 1)}$ and $co_c$ is the number of output channels of CNN, and $[;]$ denotes the concatenating operation.

After that, we obtain the sentence embeddings $Z$, which could be used during inference and in the downstream tasks.

From the perspective of the symmetry of AutoEncoder, transposed CNN is used to reconstruct the token representations. Through learning reconstruction, sentence embeddings obtain the semantics of determinative tokens. Furthermore, when calculating the similarity between two sentences, the model can find the fine-grained differences.

Specifically, we first use transposed CNN to reconstruct the multi-scales context representations:
\begin{equation}
    H'_{ks} = TransposedCNN_{ks}(Z), ks=3,4,5
\end{equation}
where the dimension of $H'_{ks}$ is same as $H_{ks}$, and the kernel size is $(3 \times 2)$.

Finally, we use the transposed TextCNN to reconstruct the tokens in sentences:
\begin{equation}
    X'_{ks} = TransposedTextCNN_{ks}(H'_{ks})
\end{equation}

And the average pooling of them are the final token representations of reconstruction:
\begin{equation}
    X' = AveragePooling(X'_3, X'_4, X'_5)
\end{equation}

Similar to the process above, we input the augmentation tokens sequence $X^+$ to get the sentence embedding $Z^+$ and reconstruction tokens $X'^+$.

\subsection{Self-Adaptive Reconstruction Loss}
Mean-square error is often used to cope with the reconstruction loss. 
However, the serious token bias of the pretrained language models towards frequency leads to the non-uniform distribution of tokens with different frequencies in representation space.
This is the reason for the vulnerability of the tokens \citep{jiang2022promptbert, 10.1145/3477495.3531823}. In this case, we propose a self-adaptive reconstruction loss based on token frequency to reduce the impact of token bias in high-frequency tokens:
\begin{gather}
    \label{eq:freq}
    f(w_i) = max(\theta, 1 - \lambda freq(w_i)) \\
    L_R = \frac{1}{N} \sum_i^N f(w_i) \times MSE(x_i, x'_i)
\end{gather}
where $\theta$ and $\lambda$ are hyper-parameters, $N$ is the number of tokens in a sentence, $freq(w_i)$ is the normalized token frequency of token $w_i$ calculated in the training set, and $MSE$ is mean-square error.

Similarly, we obtain the reconstruction loss $L_R^+$ for the augment samples:
\begin{equation}
    L_R^+ = \frac{1}{N} \sum_i^N f(w_i) \times MSE(x_i^+, {x'}_i^+)
\end{equation}

\subsection{Training Object}
Following \citet{gao-etal-2021-simcse}, contrastive learning is applied to sentence embeddings with the InfoNCE loss, which can distinguish positive samples from negative ones:
\begin{equation}
    L_I = \log \frac{e^{\text{sim}(Z_{i}, Z_{i}^+)/\tau}}{\sum^M_{j=1} e^{\text{sim}(Z_{i}, Z_{j}^+)/\tau}}
\end{equation}
where $\tau$ is a temperature hyper-parameters, $M$ is the mini-batch size, and $\text{sim}(Z_{i}, Z_{i}^+)$ is a similarity metric between $Z_{i}$ and $Z_{i}^+$. We use the cosine similarity function in this work.

Finally, the overall training object, including the contrastive loss and the reconstruction loss, is defined as:
\begin{equation}
    L = \alpha L_I + \beta L_R + \gamma L_R^+
\end{equation}
where $\alpha$, $\beta$ and $\gamma$ are hyper-parameters.

\begin{table*}[t]
\centering
\resizebox{2 \columnwidth}{!}{
\begin{tabular}{llcccccccc} 
\toprule

\textbf{Base}                  & \textbf{Model}                 & \textbf{STS12} & \textbf{STS13} & \textbf{STS14} & \textbf{STS15} & \textbf{STS16} & \textbf{STS-B} & \textbf{SICK-R} & \textbf{Avg.}   \\ 
\hline
\multirow{7}{*}{$\text{RoBERTa}_{base}$}  & first-last avg.                & 40.88          & 58.74          & 49.07          & 65.63          & 61.48          & 58.55          & 61.63           & 56.57           \\
                               & whitening \citep{su2021whitening}                      & 46.99          & 63.24          & 57.23          & 71.36          & 68.99          & 61.36          & 62.91           & 61.73           \\
                               & DeCLUTR \citep{giorgi-etal-2021-declutr}                        & 52.41          & 75.19          & 65.52          & 77.12          & 78.63          & 72.41          & 68.62           & 69.99           \\
                               & SimCSE \citep{gao-etal-2021-simcse}                         & 70.16          & \textbf{81.77} & 73.24          & 81.36          & 80.65          & 80.22          & 68.56           & 76.57           \\
                               \cdashline{2-10}
                               & SARCSE                         & \textbf{73.35} & 80.89          & \textbf{75.37} & \textbf{83.00} & \textbf{81.39} & \textbf{82.18} & \textbf{70.43}  & \textbf{78.09}  \\
                               & - w/o Self-Adaptive Loss (SAL) & 72.62          & 80.77          & 73.84          & 81.95          & 80.72          & 80.69          & 69.43           & 77.15           \\
                               & - w/o SAL \& Decoder             & 71.64          & 77.20          & 71.82          & 80.88          & 79.50          & 77.90          & 67.90           & 75.26           \\ 
\hline
\multirow{3}{*}{$\text{RoBERTa}_{large}$} & first-last avg.                & 41.24          & 55.81          & 47.99          & 64.01          & 57.81          & 53.47          & 56.95           & 53.90           \\
                               & SimCSE \citep{gao-etal-2021-simcse}                         & \textbf{72.86}          & 83.99          & 75.62          & 84.77 & \textbf{81.80} & 81.98          & 71.26           & 78.90           \\
                               & SARCSE                         & 72.28 & \textbf{84.26} & \textbf{76.81} & \textbf{84.81}          & 80.69          & \textbf{82.50} & \textbf{72.53}  & \textbf{79.13}  \\
\bottomrule
\end{tabular}
}
\caption{The results comparison with baselines on the 7 STS tasks (Spearman's correlation). The best performance is in \textbf{bold}. For more detailed experiments, please refer to the Appendix.
}
\label{tab:result}
\end{table*}

\section{Experiments}

\subsection{Experimental Setup}
\textbf{Datasets and Evaluation Metric\quad} Following the previous works, we use the 7 STS datasets as the benchmark corpus, comprised of STS tasks 2012-2016 \citep{agirre-etal-2012-semeval, agirre-etal-2013-sem, agirre-etal-2014-semeval, agirre-etal-2015-semeval, agirre-etal-2016-semeval}, STS-B \citep{cer-etal-2017-semeval} and SICK-R \citep{marelli-etal-2014-sick}. We use the development set of STS-B to choose the best model. In addition, we only use the test sets of these datasets to evaluate the model using the SentEval toolkit released by \citet{conneau2018senteval}. Similar to SimCSE, we use the 1-million sentences which are randomly sampled from English Wikipedia. And the token frequency used in the self-adaptive reconstruction loss is calculated from these sentences. For the evaluation metric, we use the Spearman's correlation coefficient between scores of cosine similarity and ground truths to get the model performance.

\noindent \textbf{Implementation Details\quad} We implement SARCSE based on Transformers\footnote{\url{https://github.com/huggingface/transformers}} \citep{wolf-etal-2020-transformers}, and use the Roberta as the pretrained language models, including the base model and large model. Additionally, the hyper-parameters of reconstruction $\theta$ and $\lambda$ are set to 0.1 and 50. The $\tau$ in InfoNCE is set to 0.05. And the $\alpha$, $\beta$ and $\gamma$ in the training object are set to 1, 2.5e-4 and 2.5e-4. The output channels of TextCNN and CNN, $co_t$ and $co_c$, are set to 500 and 3. We train SARCSE through AdamW \citep{loshchilov2018decoupled} optimizer, and the learning rate is 1e-5. Finally, we set the mini-batch to 64 and the training epoch to 1.

\subsection{Overall Results}

We show the results in Table~\ref{tab:result}. Using the $\text{RoBERTa}_{base}$ model, SARCSE shows an apparent advantage over SimCSE, especially on the STS12, STS14, STS15, STS-B and SICK-R. Additionally, SARCSE obtains a great performance improvement on average, although SARCSE does not perform well on STS13 compared with SimCSE. As for the $\text{Roberta}_{large}$ model, SARCSE still outperforms SimCSE on most tasks, although the improvement is not as huge as that on $\text{RoBERTa}_{base}$. A possible reason could be that the ability to encode sentences on large model is better than on base model. In addition, it is worth noting that SARCSE achieves this performance only by setting the batch size to 64. But the batch size of SimCSE is 512. Obviously, the process of reconstruction and self-adaptive reconstruction loss play important roles in SARCSE and reduce the dependency of SARCSE on contrastive learning.

\subsection{Ablation Study}

To further verify the effect of every component in SARCSE, we conduct the ablation studies. The results are shown in Table~\ref{tab:result}.

We first remove the self-adaptive reconstruction loss and only use the mean-square error loss, which means all the loss weights of tokens are the same. The performance slightly drops on average, demonstrating the effect of the self-adaptive reconstruction loss in reducing the influence of token bias. Moreover, SARCSE still performs better than SimCSE, which shows the effect of reconstruction by AutoEncoder and the importance of preserving the fine-grained semantics.

In addition, we remove the self-adaptive reconstruction loss and the decoder in AutoEncoder, which means the model only has a pretrained language model and a TextCNN to encode the sentences, and there is no more the process of reconstruction. The performance is even worse than SimCSE. This further illustrates the importance of reconstructing tokens in sentences. 

\section{Related Work}
Learning universal sentence embeddings is a fundamental task in NLP and has been developed for a long time. After the emergence of the pretrained language models (PLMs), most previous works directly got the sentence embeddings from them. However, \citet{li-etal-2020-sentence} found the anisotropic word embedding space in PLMs seriously impacted the performance and proposed a method which transforms the space to solve this problem. Furthermore, \citet{su2021whitening} proposed a better transformation method. From a deep learning perspective, \citet{gao-etal-2021-simcse} used contrastive learning to solve this problem and obtained the positive samples by dropout. Recently, some works proposed efficient methods to get the positive and negative samples. For example, \citet{wu-etal-2022-esimcse} got positive samples by duplicating words. \citet{jiang2022promptbert} used prompt learning to get the positive samples. On the contrary, \citet{zhou-etal-2022-debiased} and \citet{wu-etal-2022-smoothed} obtained more negative samples using Gaussian noise. In another way, SARCSE improves the model by modifying the encoder, which could be well combined with the above methods with data augmentation.

\section{Conclusion}
In this paper, we present a novel \textbf{S}elf-\textbf{A}daptive \textbf{R}econstruction \textbf{C}ontrastive \textbf{S}entence \textbf{E}mbeddings (\textbf{SARCSE}) framework by introducing a process of tokens reconstruction, which can preserve the fine-grained semantics in tokens. In addition, we propose a self-adaptive reconstruction loss based on token frequency to reduce the impact of token bias in pretrained language models. Experiments on 7 STS tasks demonstrate the effectiveness of SARCSE. 

\section*{Limitations}

In this work, we propose the self-adaptive reconstruction loss based on token frequency. However, the token frequency calculated from the training set might have sample bias, which means the self-adaptive reconstruction loss might not work well on other training sets or test sets. This can lead to a weakly robust model. We expect that a learnable reconstruction loss to be more suitable for solving the problem of token bias. We will explore it in the future.

\bibliography{anthology,custom}
\bibliographystyle{acl_natbib}

\appendix

\begin{table*}[t]
\centering
\resizebox{1.8 \columnwidth}{!}{
\begin{tabular}{clcccccccc}
\toprule
\multicolumn{1}{l}{\textbf{Batch Size}} & \textbf{Model} & \textbf{STS12} & \textbf{STS13} & \textbf{STS14} & \textbf{STS15} & \textbf{STS16} & \textbf{STS-B} & \textbf{SICK-R} & \textbf{Avg.}  \\ \hline
\multirow{2}{*}{32}                     & SimCSE         & 69.65          & \textbf{81.00} & 73.02          & \textbf{82.03} & 79.59          & \textbf{79.90} & \textbf{68.43}  & 76.23          \\
                                        & SARCSE         & \textbf{73.11} & 79.76          & \textbf{73.23} & 81.04          & \textbf{79.71} & 78.97          & 69.71           & \textbf{76.50} \\ \hline
\multirow{2}{*}{64}                     & SimCSE         & 69.12          & 80.42          & 72.02          & 81.70          & 80.20          & 79.82          & 67.32           & 75.80          \\
                                        & SARCSE         & \textbf{73.35} & \textbf{80.89} & \textbf{75.37} & \textbf{83.00} & \textbf{81.39} & 82.18          & \textbf{70.43}  & \textbf{78.09}* \\ \hline
\multirow{2}{*}{128}                    & SimCSE         & 67.18          & 80.51          & 72.03          & 81.20          & 78.95          & 78.91          & \textbf{68.48}  & 75.32          \\
                                        & SARCSE         & \textbf{72.05} & \textbf{80.59} & \textbf{73.78} & \textbf{82.13} & \textbf{80.96} & \textbf{80.59} & 68.28           & \textbf{76.91} \\ \hline
\multirow{2}{*}{192}                    & SimCSE         & 69.29          & \textbf{81.41} & 73.06          & \textbf{81.87} & \textbf{80.18} & \textbf{79.80} & \textbf{68.88}  & 76.36          \\
                                        & SARCSE         & \textbf{73.05} & 78.97          & \textbf{73.38} & 81.58          & 80.07          & 79.27          & 68.24           & \textbf{76.37} \\ \hline
\multirow{2}{*}{256}                    & SimCSE         & 68.25          & \textbf{81.53} & \textbf{73.76} & \textbf{82.34} & \textbf{80.76} & 80.35          & \textbf{68.88}  & 76.55          \\
                                        & SARCSE         & \textbf{72.51} & 79.89          & 73.49          & 81.86          & \textbf{80.76} & \textbf{80.51} & 68.32           & \textbf{76.76} \\ \hline
\multirow{2}{*}{512}                    & SimCSE         & 70.16          & \textbf{81.77} & \textbf{73.24} & 81.36          & \textbf{80.65} & 80.22          & 68.56           & 76.57*          \\
                                        & SARCSE         & \textbf{73.12} & 79.47          & 73.16          & \textbf{81.53} & 80.48          & \textbf{80.62} & \textbf{68.69}  & \textbf{76.72} \\ 
\bottomrule
\end{tabular}
}
\caption{The results comparison with SimCSE on the 7 STS tasks by setting different batch sizes. The results with * mean the best performance of the corresponding model across all batch sizes. All models are based on $\text{RoBERTa}_{base}$.}
\label{tab:batch size}
\end{table*}

\begin{table*}[t]
\centering
\resizebox{1.6 \columnwidth}{!}{
\begin{tabular}{ccccccccc}
\toprule
\textbf{$\theta$} & \textbf{STS12} & \textbf{STS13} & \textbf{STS14} & \textbf{STS15} & \textbf{STS16} & \textbf{STS-B} & \textbf{SICK-R} & \textbf{Avg.}  \\ \hline
0              & 73.05          & \textbf{81.24} & 73.89          & 82.07          & 81.19          & 81.64          & 68.90           & 77.43          \\
0.1            & 73.35          & 80.89          & \textbf{75.37} & \textbf{83.00} & \textbf{81.39} & \textbf{82.18} & \textbf{70.43}  & \textbf{78.09} \\
0.2            & 71.61          & 79.01          & 73.24          & 81.76          & 80.60          & 80.49          & 69.08           & 76.54          \\
0.3            & \textbf{74.71} & 80.15          & 73.18          & 81.91          & 80.00          & 80.22          & 69.33           & 77.07          \\
0.4            & 73.22          & 80.91          & 73.65          & 82.01          & 80.84          & 81.11          & 68.70           & 77.21          \\
0.5            & 71.51          & 80.87          & 73.91          & 82.42          & 80.94          & 81.05          & 69.44           & 77.16          \\
0.6            & 72.16          & 80.33          & 73.92          & 82.16          & 80.61          & 80.84          & 69.58           & 77.09          \\
\bottomrule
\end{tabular}
}
\caption{The results of SARCSE on the 7 STS tasks by setting different $\theta$. All models are based on $\text{RoBERTa}_{base}$.}
\label{tab:theta}
\end{table*}

\section{The Impact of Batch Size}

It is well-known that the batch size is essential in contrastive learning. The bigger batch size means that more negative pairs could be constructed, which can further improve the performance of models. To this end, we explore the impact of batch size. As shown in Table~\ref{tab:batch size}, SARCSE outperforms SimCSE in the case of all batch sizes, which shows the effectiveness and robustness of SARCSE. 

Besides, SimCSE reaches the best performance with the batch size of 512. The experimental results illustrate that SimCSE is highly dependent on a large number of negative samples in contrastive learning. This further results in the SimCSE requiring more computational resources to train the model. For example, SimCSE needs 2 NVIDIA GeForce RTX 3090 24GB when the batch size is 512. On the contrary, SARCSE achieves the best performance when the batch size is only 64, which means the process of self-adaptive reconstruction can help the model to be optimized in another direction and reduce the reliance on contrastive learning. Finally, the small batch size can reduce the requirement for computational resources. SARCSE can even be trained on a single NVIDIA GeForce GTX 1080 Ti 11GB and reach the best performance.

\section{The Impact of Hyper-Parameter $\theta$}

In Equation~\ref{eq:freq}, we use a hyper-parameter $\theta$ to limit the minimum weight of loss for each token reconstruction. To quantitatively study its effect on the model performance, we experiment by setting $\theta$ from 0 to 0.6 with each increase by 0.1. The results are shown in Table~\ref{tab:theta}. SARCSE achieves the best performance when $\theta$ is 0.1. When $\theta$ is set to 0, some tokens with high frequency will not be learned in reconstruction. This result in a breakdown of semantic coherence and further degrades the performance. On the other hand, when $\theta$ is increased, more tokens will reach the minimum weight, which means the different importance of tokens in the loss function is disappeared. And the ability of the self-adaptive reconstruction loss to solve the token bias problem is weakened. Finally, it further hurts the performance. We argue setting $\theta$ to 0.1 is perfectly balanced and reasonable. In this setting, only some extremely common tokens (e.g., "the", "of", and some punctuation) will have their weight decreased to 0.1. Most tokens retain their distinction in weight, further improving the performance of SARCSE.

\section{Alignment and Uniformity}
\label{sec: alignment}

Following \citet{wang2020understanding}, we utilize alignment and uniformity to evaluate the quality of sentence embeddings. Given a distribution of positive sentence pairs $p_{\text{pos}}$ which are similar and a distribution of sentences in whole dataset $p_{\text{data}}$. Alignment and uniformity are defined as:

\begin{gather}
    \ell_{\text{align}} \triangleq \mathop{\mathbb{E}}\limits_{(x, x^{+}) \sim p_{\text{pos}}} {\Vert f(x)-f(x^+) \Vert} ^2 \\
    \ell_{\text{uniform}} \triangleq \text{log} \mathop{\mathbb{E}}\limits_{\mathop{_{i.i.d.}}\limits_{x,y \sim p_{\text{data}}}} e^{-2 {\Vert f(x)-f(y) \Vert}^2}
\end{gather}

\begin{table}[t!]
\centering
\resizebox{1 \columnwidth}{!}{
\begin{tabular}{lccc}
\toprule
\multicolumn{1}{c}{\textbf{Model}} & \textbf{Alignment} & \textbf{Uniformity} & \textbf{STS Avg.} \\ \hline
firts-last avg.                    & \textbf{0.021}     & -0.215              & 56.57             \\
SimCSE                             & 0.213              & \textbf{-2.618}     & 76.57             \\
SARCSE                             & 0.148              & -2.213              & \textbf{78.09}   \\
\bottomrule
\end{tabular}
}
\caption{The alignment and uniformity of SARCSE and baselines. All models are based on $\text{RoBERTa}_{base}$.}
\label{tab:alignment}
\end{table}

The alignment measures the distance between positive pairs. And the distance should be low in positive pairs. The uniformity measures the sentence embeddings distribution in the representation space. And the distance should be high in random pairs. Thus, the smaller $\ell_{\text{align}}$ and $\ell_{\text{uniform}}$ are better.

We show the results of alignment and uniformity in Table~\ref{tab:alignment}. The RoBERTa-base first-last Avg. gets the best alignment but the worst uniformity because of the well-known high anisotropy in pretrained language models. To this end, SimCSE proposed to solve this anisotropy problem by contrastive learning. However, although the uniformity is better in SimCSE, the alignment is not learned in training, which makes it sub-optimal and neglected. We argue the reconstruction in SARCSE helps the model catch the fine-grained semantics and further improves the alignment in feature representations with only a slight decrease in uniformity. Finally, SARCSE improves the performance on 7 STS tasks by balancing the alignment and uniformity.

\begin{figure}[t!]
    \centering
    \subfigure[SimCSE]{\includegraphics[width=\linewidth]{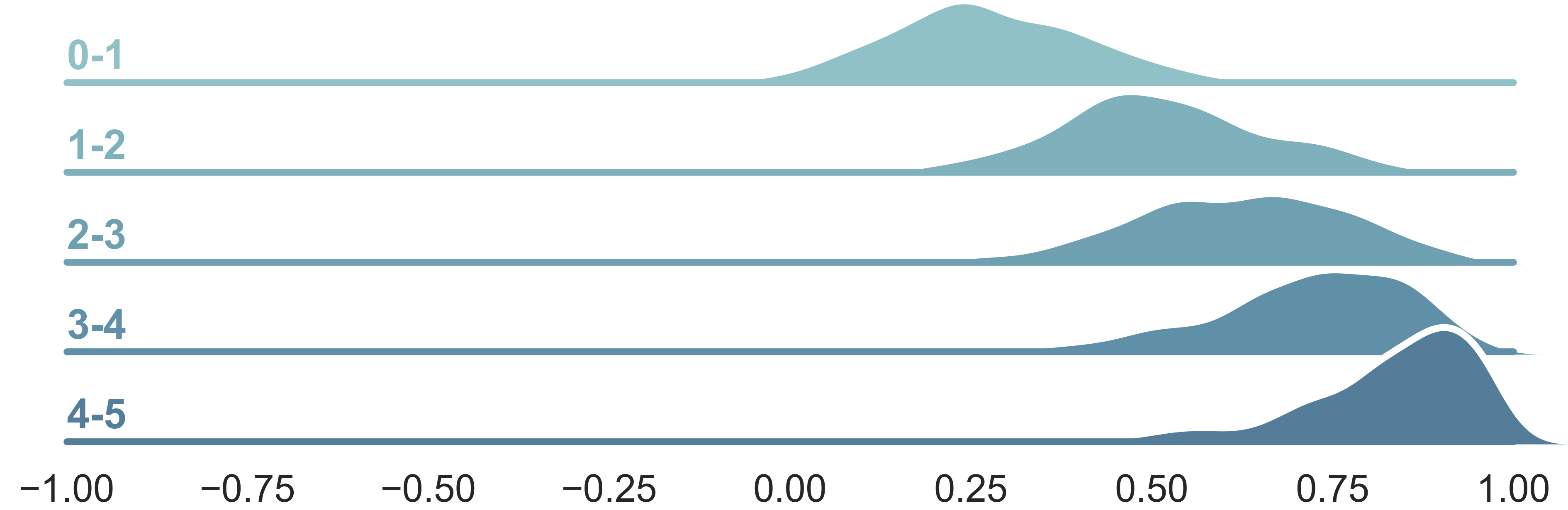}}
    \subfigure[SARCSE]{\includegraphics[width=\linewidth]{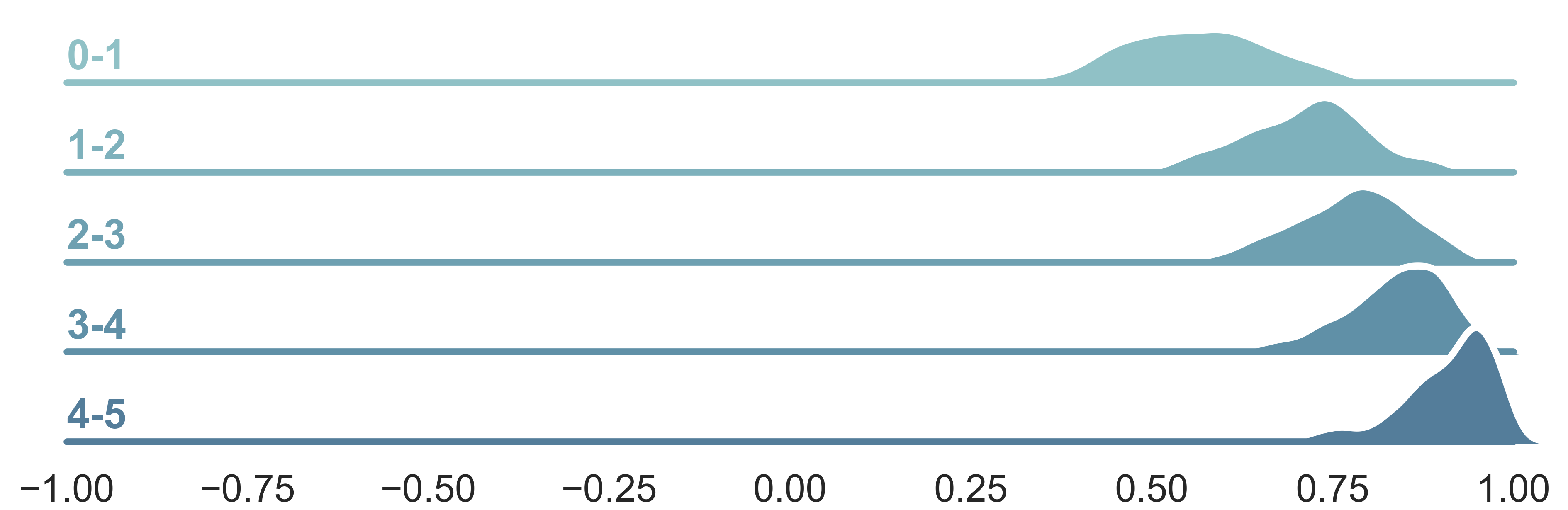}}
    \caption{
    The density plots of SimCSE and SARCSE in the test set of STS-B. The data are divided into 5 groups bu the ground truth similarity ratings. The higher ratings mean more similar. The y-axis represents the grouping situation, while the x-axis represents the cosine similarity.
    }
    \label{fig:sim}
\end{figure}

\section{Cosine-Similarity Distribution}
To further show the differences between SARCSE and SimCSE, we visualize their density plots of the predicted results in Figure~\ref{fig:sim}. The test set of STS-B is divided into five groups (i.e., 0-1, 1-2, 2-3, 3-4, and 4-5) by the ground truth similarity ratings. And we build the density plots for each group. Compared with SimCSE, the predictions of SARCSE has lower variance in each group. Although the peak values of cosine similarities in the five groups became closer, SARCSE still achieves better performances. Obviously, the lower variance is more important in the metric of Spearman's correlation. Furthermore, the density plots also explain the differences in alignment and uniformity between SimCSE and SARCSE, which we have shown in Appendix~\ref{sec: alignment}. SARCSE obtain a better alignment, which means the positive examples are more similar. Hence, the density plots of predictions in SARCSE have a lower variance.

\end{document}